\documentclass[11pt]{article}
\usepackage{eacl2017}
\usepackage{times}
\usepackage{url}
\usepackage{latexsym}
\usepackage{booktabs}
\usepackage{rotating}
\usepackage{todonotes}
\usepackage{color}
\usepackage{amsmath}
\usepackage{mathtools,xparse}
\usepackage{multirow}
\usepackage{xcolor}

\eaclfinalcopy 
\def\eaclpaperid{337}

\title{Re-evaluating Automatic Metrics for Image Captioning}
\author{Mert Kilickaya, Aykut Erdem, Nazli Ikizler-Cinbis, and Erkut Erdem \\
  Hacettepe University Computer Vision Lab\\Dept. of Computer Engineering, Hacettepe University, Ankara, TURKEY \\
  {\tt \small kilickayamert@gmail.com,\{aykut,nazli,erkut\}@cs.hacettepe.edu.tr}}
\date{}

\begin{document}
\maketitle
\begin{abstract}
The task of generating natural language descriptions from images has received a lot of attention in recent years. Consequently, it is becoming increasingly important to evaluate such image captioning approaches in an automatic manner. In this paper, we provide an in-depth evaluation of the existing image captioning metrics through a series of carefully designed experiments. Moreover, we explore the utilization of the recently proposed Word Mover's Distance (\textsc{wmd}) document metric for the purpose of image captioning. Our findings outline the differences and/or similarities between metrics and their relative robustness by means of extensive correlation, accuracy and distraction based evaluations. Our results also demonstrate that \textsc{wmd} provides strong advantages over other metrics.
\end{abstract}

\section{Introduction}
\label{sec:introduction}

There has been a growing interest in research on integrating vision and language in  natural language processing and computer vision communities. As one of the key problems in this emerging area, image captioning aims at generating natural descriptions of a given image~\cite{bernardi2016automatic}. This is a challenging problem since it requires the ability to not only understand the visual content, but also to generate a linguistic description of that content. In this regard, it can be framed as a machine translation task where the source language denotes the visual domain and the target language is a specific language such as English. The recently proposed deep image captioning studies follow this interpretation and model the process via an encoder-decoder architecture~\cite{vinyals2015show,xu2015show,karpathy2015deep,jia2015guiding}. These approaches have attained considerable success in the recent benchmarks such as \textsc{flickr8k}~\cite{hodosh2013framing}, \textsc{flickr30k}~\cite{young2014image} and \textsc{ms coco}~\cite{lin2014microsoft} as compared to the earlier techniques which explicitly detect objects and generate descriptions by using surface realization techniques~\cite{kulkarni2013babytalk,li2011composing,elliott2013image}. 

With the size of the benchmark datasets becoming larger and larger, evaluating image captioning models has become increasingly important. Human-based evaluations become obsolete as they are costly to acquire and, more importantly, not repeatable. Automatic evaluation metrics are employed as an alternative to human evaluation in both developing new models and comparing them against the state-of-the-art. These metrics compute a score that indicates the similarity/dissimilarity between an automatically generated caption and a number of human-written reference (gold standard) descriptions. 

 \begin{table*}[t]
  \centering
  \caption{A summary of the evaluation metrics considered in this study.}
  \resizebox{\linewidth}{!}{
  \begin{tabular}{lll}
  \toprule 
Metric & Proposed to evaluate & Underlying idea\\
\midrule
\textsc{bleu}~\cite{papineni2002bleu} & Machine translation & $n$-gram precision\\
\textsc{rouge}~\cite{lin2004rouge} & Document summarization & $n$-gram recall\\
\textsc{meteor}~\cite{banerjee2005meteor} & Machine translation  & $n$-gram with synonym matching\\
\textsc{cide}r~\cite{vedantam2015cider} & Image description generation & {\em tf-idf} weighted $n$-gram similarity\\
\textsc{spice}~\cite{spice2016} & Image description generation  & Scene-graph synonym matching\\
\textsc{wmd}~\cite{kusner2015word} & Document similarity & Earth Mover Distance on {\em word2vec}\\
\bottomrule  
  \end{tabular}
  } 
   \label{tab:metrics}
\end{table*}

Some of these automatic metrics such as \textsc{bleu}~\cite{papineni2002bleu}, \textsc{rouge}~\cite{lin2004rouge},  \textsc{meteor}~\cite{banerjee2005meteor}, and \textsc{TER}~\cite{snover2006ter} have originated from the readily available metrics for machine translation and/or text summarization. On the contrary, the more recent metrics such as \textsc{cide}r~\cite{vedantam2015cider} and \textsc{spice}~\cite{spice2016}  are specifically developed for image caption evaluation task. 

Evaluation with automatic metrics has some challenges as well. As previously analyzed in~\cite{elliott-keller:2014:P14-2}, the existing automatic evaluation measures have proven to be inadequate in successfully mimicking the human judgements for evaluating the image descriptions. The latest evaluation results of 2015 \textsc{ms~coco} Challenge on image captioning has also revealed some interesting findings in line with this observation~\cite{vinyals2016pami}. In the challenge, the recent deep models outperform the human upper bound according to automatic measures, yet they could not beat the humans when the subjective human judgements are considered. These demonstrate that we need to better understand the drawbacks of existing automatic evaluation metrics. This motivates us to present an in-depth analysis of the current metrics employed in image description evaluation. 

We first review \textsc{bleu}, \textsc{rouge}, \textsc{meteor}, \textsc{cide}r and \textsc{spice} metrics, and discuss their main drawbacks. In this context, we additionally describe \textsc{wmd} metric which has been recently proposed as a distance measure between text documents in~\cite{kusner2015word}. We then investigate the performance of these automatic metrics through different experiments. We analyze how well these metrics mimic human assessments by estimating their correlations with the collected human judgements. Different from the previous related work~\cite{elliott-keller:2014:P14-2,vedantam2015cider,spice2016}, we perform a more accurate analysis by additionally reporting the results of Williams significance test. This further allows us to figure out the differences and/or similarities between a pair of metrics, whether any two metrics complement each other or provide similar results. We then test the ability of these metrics to distinguish certain pairs of captions from one another in reference to a ground truth caption. Next, we carry out an analysis on  robustness of these metrics by analyzing how well they cope with the distractions in the descriptions~\cite{hockenmaier-acl-ws}. 

\section{Evaluation Metrics}
\label{sec:metrics}

A summary of the metrics investigated in our study is given in Table~\ref{tab:metrics}. All these metrics except \textsc{spice} and \textsc{wmd} define the similarity over words or $n$-grams of reference and candidate descriptions by considering different formulas. On the other hand, \textsc{spice}~\cite{spice2016} considers a scene-graph representation of an image by encoding objects, their attributes and relations between them, and \textsc{wmd} leverages word embeddings to match groundtruth descriptions with generated captions.

\subsection{\textsc{bleu}}
\label{ssec:bleu}
\noindent \textsc{bleu}~\cite{papineni2002bleu} is one of the first metrics that have been in use for measuring similarity between two sentences. It has been initially proposed for machine translation, and defined as the geometric mean of $n$-gram precision scores multiplied by a brevity penalty for short sentences. In our experiments, we use the smoothed version of \textsc{bleu} as described in ~\cite{lin-och:2004:ACL}.

\subsection{\textsc{rouge}}
\label{ssec:rouge}
\noindent \textsc{rouge}~\cite{lin2004rouge} is initially proposed for evaluation of summarization systems, and this evaluation is done via comparing overlapping $n$-grams, word sequences and word pairs. In this study, we use \textsc{rouge-l} version, which basically measures the longest common subsequences between a pair of sentences. Since \textsc{rouge} metric relies highly on recall, it favors long sentences, as also noted by~\cite{vedantam2015cider}.

\subsection{\textsc{meteor}}
\label{ssec:meteor}
\noindent \textsc{meteor}~\cite{banerjee2005meteor} is another machine translation metric. It is defined as the harmonic mean of precision and recall of uni-gram matches between sentences. Additionally, it makes use of synonyms and paraphrase matching. \textsc{meteor} addresses several deficiencies of \textsc{bleu} such as recall evaluation and the lack of explicit word matching. $n$-gram based measures work reasonably well when there is a significant overlap between reference and candidate sentences; however they fail to spot semantic similarity when the common words are scarce. \textsc{meteor} handles this issue to some extent using WordNet-based synonym matching, however just looking at synonyms may be too restrictive to capture overall semantic similarity. 

\subsection{\textsc{cide}r}
\label{ssec:cider}
\noindent \textsc{cide}r~\cite{vedantam2015cider} is a recent metric proposed for evaluating the quality of image descriptions. It measures the consensus between candidate image description $c_i$ and the reference sentences, which is a set $S_i= \{s_{i1},\ldots,s_{im}\}$ provided by human annotators. For calculating this metric, an initial stemming is applied and each sentence is represented with a set of 1-4 grams. Then, the co-occurrences of $n$-grams in the reference sentences and candidate sentence are calculated. In \textsc{cide}r, similar to {\em tf-idf}, the $n$-grams that are common in all image descriptions are down-weighted. Finally, the cosine similarity between $n$-grams (referred as \textsc{cide}r$_n$) of the candidate and the references is computed.

\textsc{cide}r is designed as a specialized metric for image captioning evaluation, however, it works in a purely linguistic manner, and only extends existing metrics with {\em tf-idf} weighting over $n$-grams. This sometimes causes unimportant details of a sentence to be weighted more, resulting in a relatively ineffective caption evaluation. 

\subsection{\textsc{spice}}
\label{ssec:spice}
\begin{figure}[!t]
\centering
\includegraphics[width=0.495\textwidth]{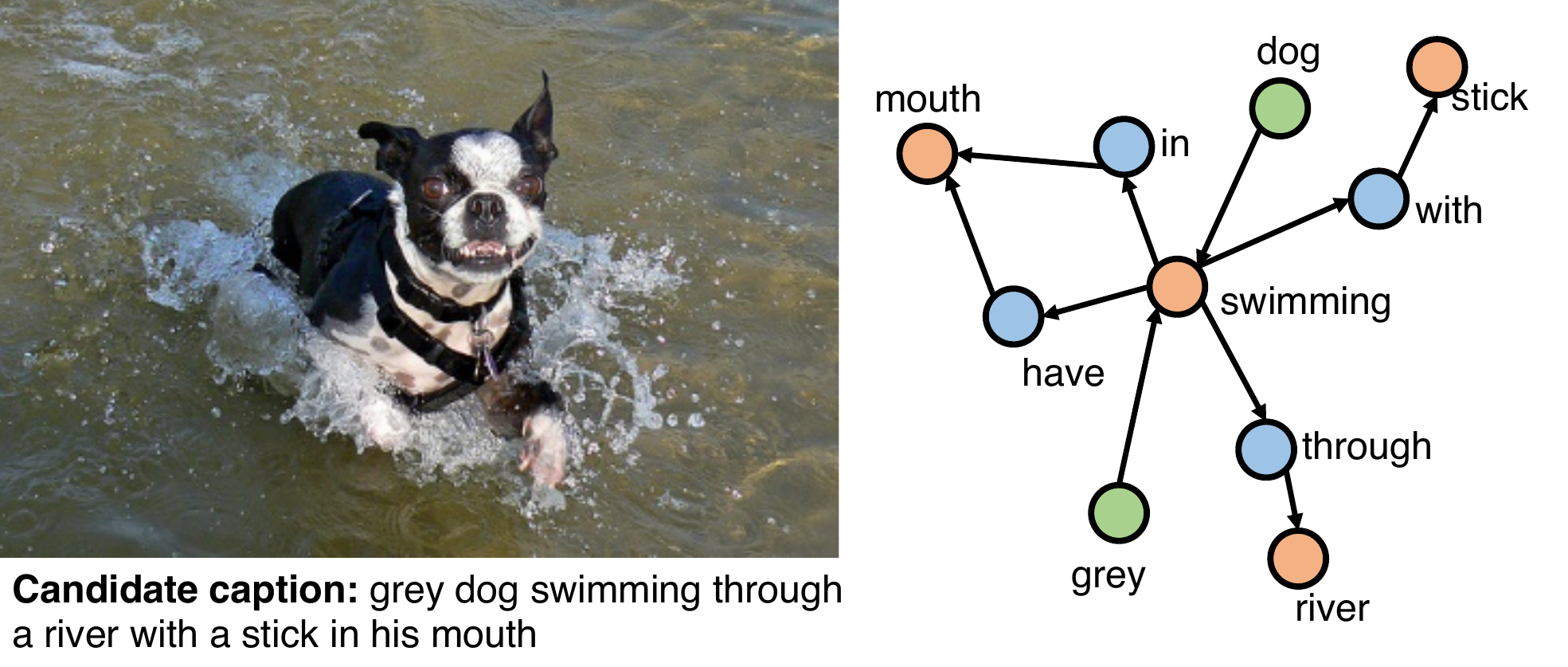}
\caption{An example image with its Scene Graph where the parser fails to parse the candidate sentence accurately, which could result in wrong calculation of \textsc{spice} metric. See text for details.}
\label{fig:scenegraph}
\end{figure}

\noindent Another recently proposed metric for evaluating image caption similarity is \textsc{spice}~\cite{spice2016}. It is based on the agreement of the scene-graph tuples~\cite{johnson2015image,schuster2015generating} of the candidate sentence and all reference sentences. Scene-graph is essentially a semantic representation that parses the given sentence to semantic tokens such as object classes $C$, relation types $R$ and attribute types $A$. Formally, a candidate caption $c$ is parsed into a scene-graph as 
\[
G(c) = \langle O(c), E(c), K(c) \rangle
\]

\noindent where $G(c)$ denotes the scene graph of caption $c$, $O(c) \subseteq C$ is the set of object mentions, $E(c) \subseteq {O(c) \times R \times O(c)}$ is the set of hyper-edges representing relations between objects, and $K(c) \subseteq O(c) \times A$ is the set of attributes associated with objects. Once the parsing is done, a set of tuples is formed by using the elements of $G$ and their possible combinations. \textsc{spice} score is then defined as the $F_{1}$-score based on the agreement between the candidate and reference caption tuples. For tuple matching, \textsc{spice} uses WordNet synonym matching~\cite{pedersen2004wordnet} as in \textsc{meteor}~\cite{banerjee2005meteor}. One problem is that the performance becomes quite dependent on the quality of the parsing. Figure~\ref{fig:scenegraph} illustrates an example case of failure. Here, \textsf{\small{swimming}} is parsed as an object, with all its relations, and \textsf{\small{dog}} is parsed as an attribute.

\subsection{\textsc{wmd}}
\label{ssec:wmd}

Two captions may not share the same words or any synonyms; yet they can be semantically similar. On the contrary, two captions may include similar objects, attributes or relations yet they may not be semantically similar. Metrics that are currently in use fail to correctly identify and assess the quality of such cases. To address this issue, we propose to use a recently introduced document distance measure called Word Mover's Distance (\textsc{wmd})~\cite{kusner2015word} for evaluating image captioning approaches. \textsc{wmd} casts the distance between documents as an instance of Earth Mover's Distance (\textsc{emd})~\cite{rubner2000earth}, where travel costs are calculated based on {\em word2vec}~\cite{mikolov2013efficient} embeddings of the words.

 \begin{table*}[!t]
  \centering
  \caption{Drawbacks of automatic evaluation metrics for image captioning. See text for details.}
\resizebox{\textwidth}{!}{
  \begin{tabular}{p{2cm}p{7cm}cccccc}
\toprule
 & Description & \textsc{bleu} & \textsc{meteor} & \textsc{rouge} & \textsc{cide}r & \textsc{spice} & \textsc{wmd} \\ \midrule 
original & \small{\textsf{a man wearing a red life jacket is sitting in a canoe on a lake}} &
1 &  1 &  1 &  10  & 1 &  1 \\  

candidate & \small{\textsf{a man wearing a life jacket is in a small boat on a lake}} &
0.45 &  0.28 &  0.67 &  2.19  & 0.40 &  0.19 \\   \midrule

synonyms & \small{\textsf{a \textbf{guy} wearing a life \textbf{vest} is in a small boat on a lake}} &
\textbf{0.20} &  \textbf{0.17} &  0.57 &  \textbf{0.65}  & \textbf{0.00} &  \textbf{0.10} \\  

redundancy & \small{\textsf{a man wearing a life jacket is in a small boat on a lake \textbf{at sunset}}} &
0.45 &  0.28 &  0.66 &  2.01  & 0.36 &  0.18 \\   

word order & \small{\textsf{\textbf{in a small boat on a lake} a man is wearing a life jacket}}  &
\textbf{0.26} &  0.26 &  \textbf{0.38} &  \textbf{1.32}  & 0.40 &  0.19 \\  \bottomrule 
               
  \end{tabular}  }
   \label{tab:hand-distractor}
\end{table*}

For \textsc{wmd}, text documents (in our case image captions) are first represented by their normalized bag-of-words (n\textsc{bow}) vectors, accounting for all words except stopwords. More formally, each text document is represented as vectors $\mathbf{d}\in R^n$, where, $d_i = \frac{c_i}{\Sigma_{j=1}^n c_j}$ if a word $i$ appears $c_i$ times in the document. \textsc{wmd} incorporates semantic similarity between individual word pairs into the document similarity metric, by using the distances in {\em word2vec} embedding space. Specifically, the distance between word $i$ and word $j$ in two documents is set as the Euclidean distance between each of the corresponding {\em word2vec} embeddings $x_i$ and $x_j$, {\em i.e.}, $c(i,j) = {\| x_i - x_j \|}_2$. 

\begin{figure}[!t]
\centering
\includegraphics[width=0.5\textwidth]{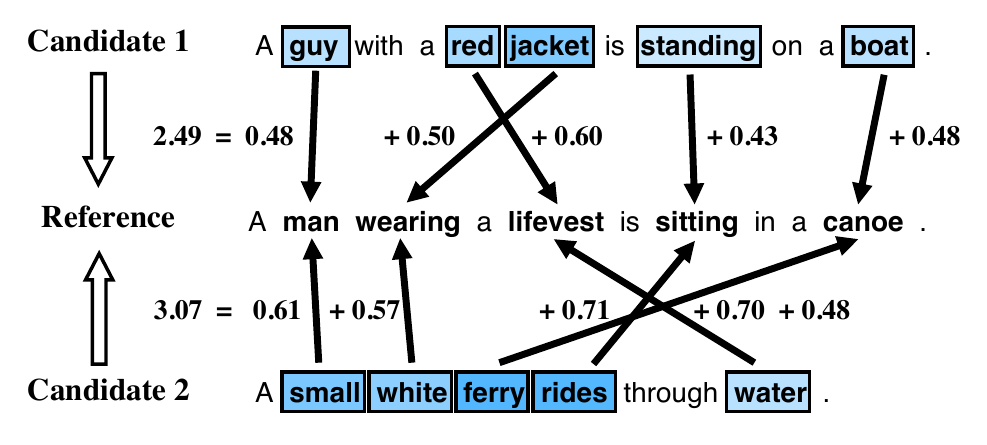}
\caption{An illustration of the distance calculation of \textsc{wmd} metric comparing two candidate captions with a reference caption.}
\label{fig:wmd}
\end{figure}

The distances between words serve as building blocks to define distances between documents, hence captions. The flow between word vectors is defined with the sparse flow matrix $\mathbf{T} \in R^{n \times n}$, with $T_{ij}$ representing the travel  amount of word $i$ to word $j$. The distance between two documents is then defined with $\Sigma_{i,j}{{T_{ij}}c(i,j)}$, {\em i.e.} the minimum cumulative cost required to move all words between documents. This minimum cumulative cost is found by solving the corresponding linear optimization problem, which is cast as a special case of \textsc{emd} metric~\cite{rubner2000earth}. An example matching result is shown in Figure~\ref{fig:wmd}. By using {\em word2vec} embeddings, semantic similarities between words are more accurately identified. In our experiments, we convert the distance scores to similarities by using a negative exponential.

\subsection{Drawbacks of the metrics}
In order to illustrate the drawbacks of these automatic evaluation metrics, we provide an example case in Table \ref{tab:hand-distractor}. In this table, an original caption is given, together with the upper bound values for each metric, {\em i.e.} when this original caption is compared to itself. The second line includes a candidate caption that is semantically very similar to the original one and the corresponding similarity scores according to evaluation metrics. We then modify the candidate sentence slightly and observe how the metric scores are affected from these small modifications. First, we observe that all the scores decrease when some words are replaced with their synonyms. The change is especially significant for \textsc{spice} and \textsc{cide}r. In this example, failure of \textsc{spice} is likely due to incorrect parsing or the failure of synonym matching. On the other hand, failure of \textsc{cide}r is likely due to unbalanced {\em tf-idf} weighting. Second, we observe that the metrics are not affected much from the introduction of additional (redundant) words in the sentences. However, when the order of the words are changed, we see that \textsc{bleu}, \textsc{rouge} and \textsc{cide}r scores decrease notably, due to their dependence on $n$-gram matching. Note that, \textsc{wmd} and \textsc{spice} are not influenced from the change in word order.

\section{Evaluation and Discussion}
\label{sec:evaluation}
\subsection{Quality}
\label{ssec:williams}

A common way of assessing the performance of a new automatic image captioning metric is to analyze how well it correlates with human judgements of description quality. However, in the literature, there is no consensus on which correlation coefficient is best suited for measuring the soundness of a metric in this way.  Elliott and Keller~\shortcite{elliott-keller:2014:P14-2} reports Spearman's rank correlation, which measures a monotonic relation, whereas Anderson \textit{et al.} \shortcite{spice2016} suggests to use Pearson's correlation, which assumes that the relation is linear, and Kendall's correlation, which is another rank correlation measure.

The above correlation analysis is a well-established practice for automatic metric evaluation, but it is not complete in the sense that it is not meaningful to draw conclusions from it about the differences or similarities between a pair of metrics. That is, comparing the corresponding correlations relative to each other does not say much since they are both computed on the same dataset, and thus not independent. To address this issue, Graham and Baldwin~\shortcite{graham-baldwin:2014:EMNLP2014} have suggested to use Williams significance test~\cite{williams1959}, which also takes into account the degree to which the two metrics correlate with each other, and can reveal whether one metric significantly outperforms the other. The test has shown to be valuable for evaluation of document and segment-level machine translation~\cite{graham-baldwin:2014:EMNLP2014,grahametal:15,graham2016accurate} and summarization metrics~\cite{graham:2015:EMNLP}. In this study, we extend the  previous correlation-based evaluations of image captioning metrics by providing a more conclusive analysis based on Williams significance test.

Williams test~\cite{williams1959} calculates the statistical significance of differences in dependent correlations, and formulated as testing whether the population correlation between $X_1$ and $X_3$ equals the population correlation between $X_2$ and $X_3$: 
\begin{equation}
t(n-3) = \frac{(r_{13} - r_{23})\sqrt{(n-1)(1+r_{12})}}{\sqrt{2K(\frac{n-1}{n-3}) + \frac{(r_{23}+r_{13})^2}{4}(1-r_{12})^3}}
\end{equation} 

\noindent where $r_{ij}$ is the correlation between $X_i$ and $X_j$, and $n$ is the size of the population, with
\begin{equation}
K = 1 - r_{12}^2 - r_{13}^2 - r_{23}^2 + 2r_{12}r_{13}r_{23}.
\end{equation}

\begin{table*}[!t]
\centering
\caption{Correlation between automatic image captioning metrics and human judgement scores.}
\resizebox{0.8\linewidth}{!}{
\begin{tabular}{l@{\hspace{0.05\linewidth}}rrr@{\hspace{0.075\linewidth}}rrr} \toprule
& \multicolumn{3}{c}{\textsc{flickr-8k}} & \multicolumn{3}{c}{\textsc{composite}} \\ 
 & Pearson & Spearman & Kendall & Pearson & Spearman & Kendall \\\midrule  
\textsc{wmd} 	  & 0.68 & 0.60 & 0.48 & \textbf{0.43} & 0.43 & 0.32 \\ 
\textsc{spice}    & \textbf{0.69} & \textbf{0.64} & \textbf{0.56} & 0.40 & 0.42 & \textbf{0.34}\\ 
\textsc{cide}r    & 0.60	 & 0.56 & 0.45 & 0.32 & 0.42 & 0.32\\ 
\textsc{meteor}   & \textbf{0.69} & 0.58 & 0.47 & 0.37 & \textbf{0.44} & 0.33\\ 
\textsc{bleu}     & 0.59 & 0.44 & 0.35 & 0.34 & 0.38 & 0.28\\ 
\textsc{rouge}    & 0.57 & 0.44 & 0.35 & 0.40 & 0.39 & 0.29\\  
\bottomrule 
\end{tabular}
}
\label{tab:correlations}
\end{table*}

\begin{figure*}[!t]
\centering
\includegraphics[width=0.32\textwidth]{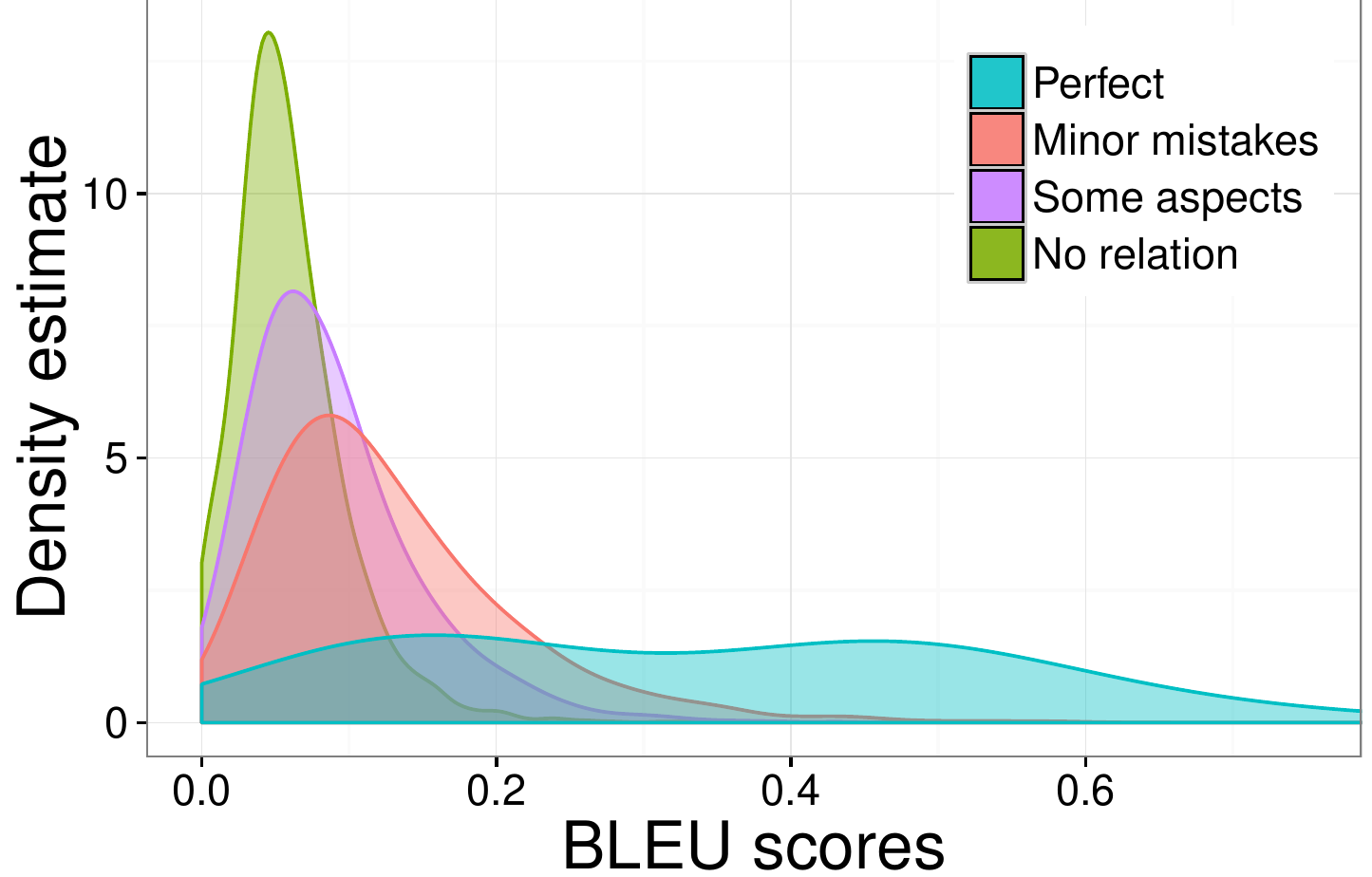}
\includegraphics[width=0.32\textwidth]{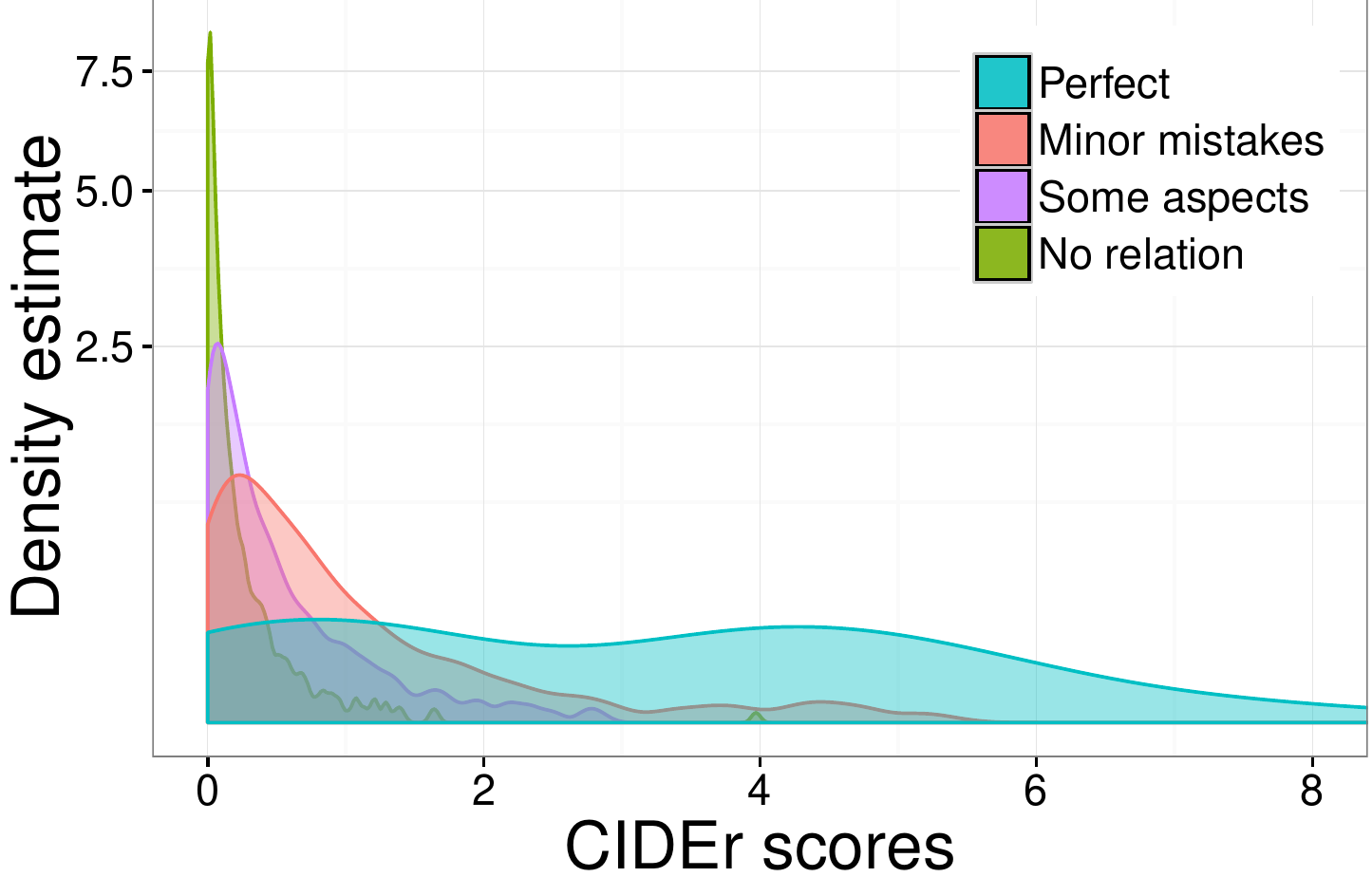}
\includegraphics[width=0.32\textwidth]{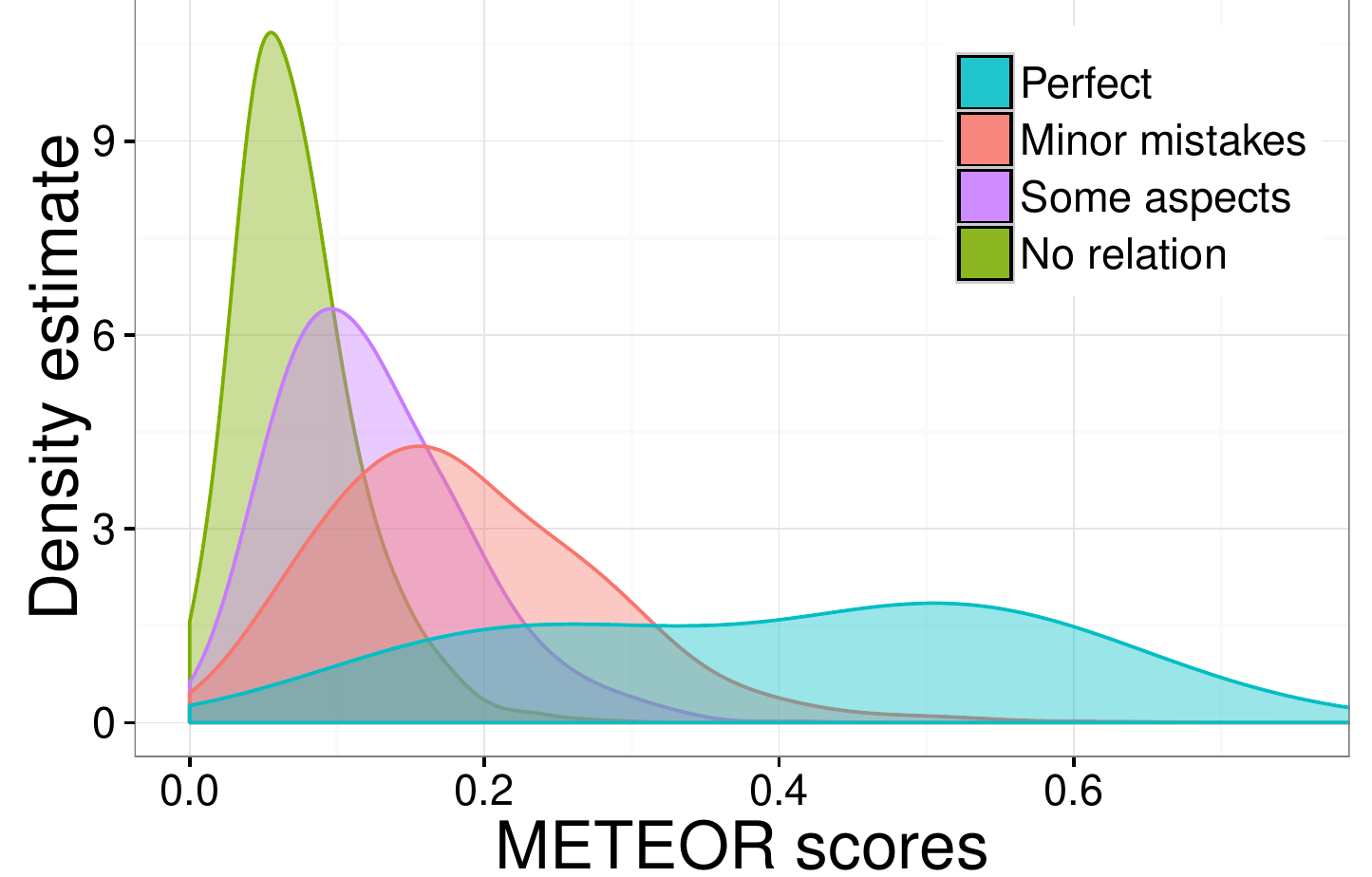}
\includegraphics[width=0.32\textwidth]{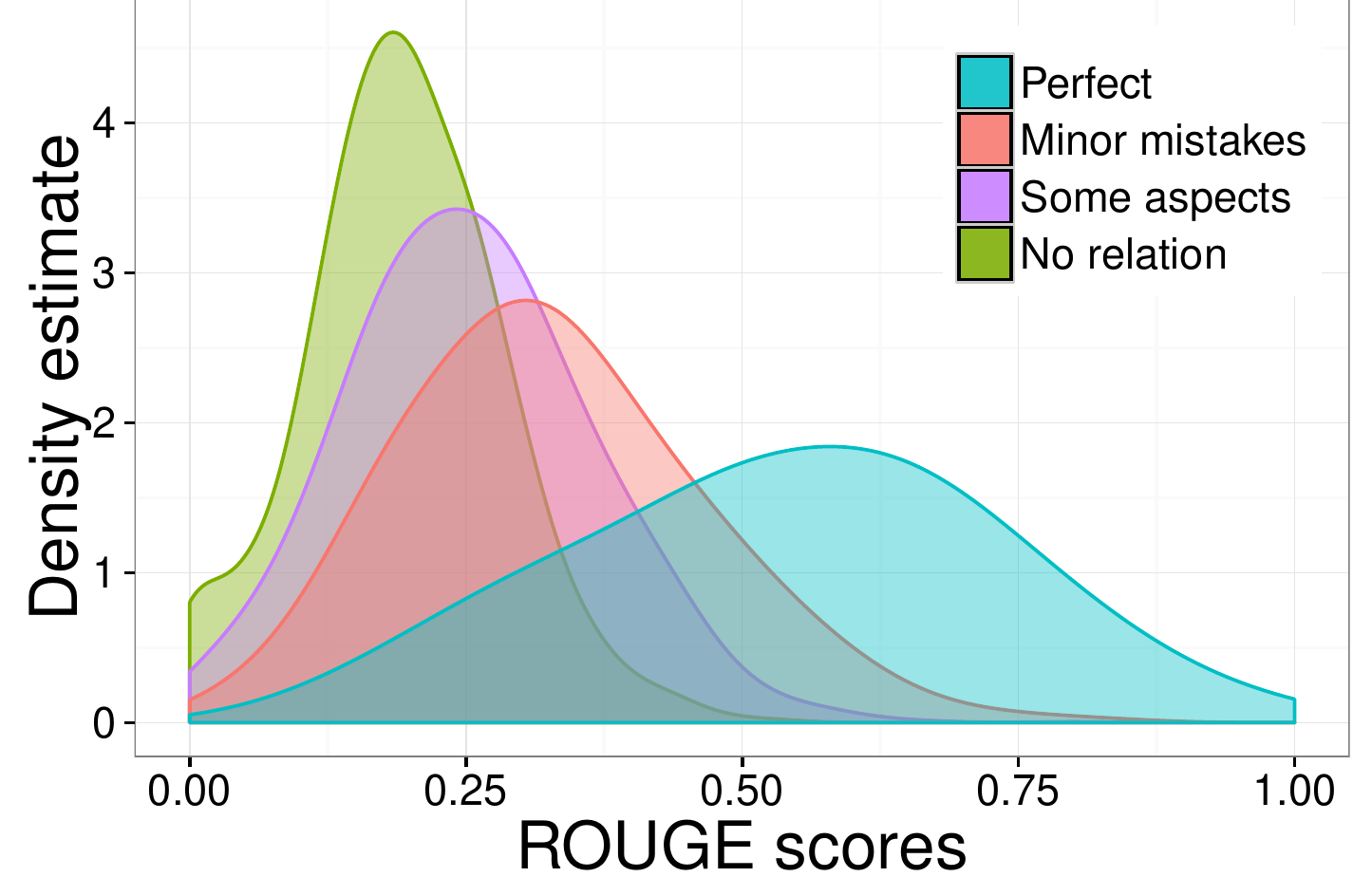}
\includegraphics[width=0.32\textwidth]{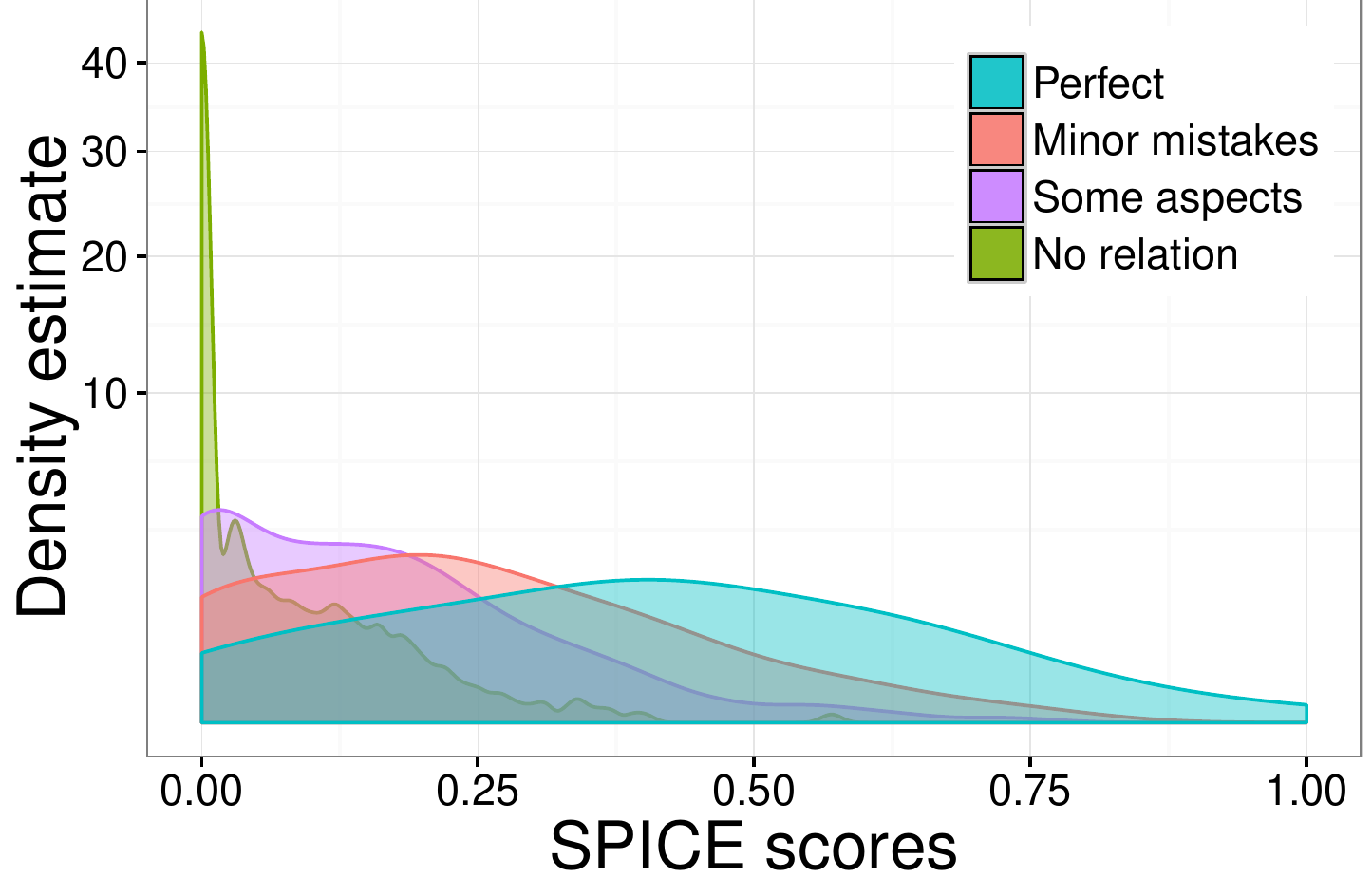}
\includegraphics[width=0.32\textwidth]{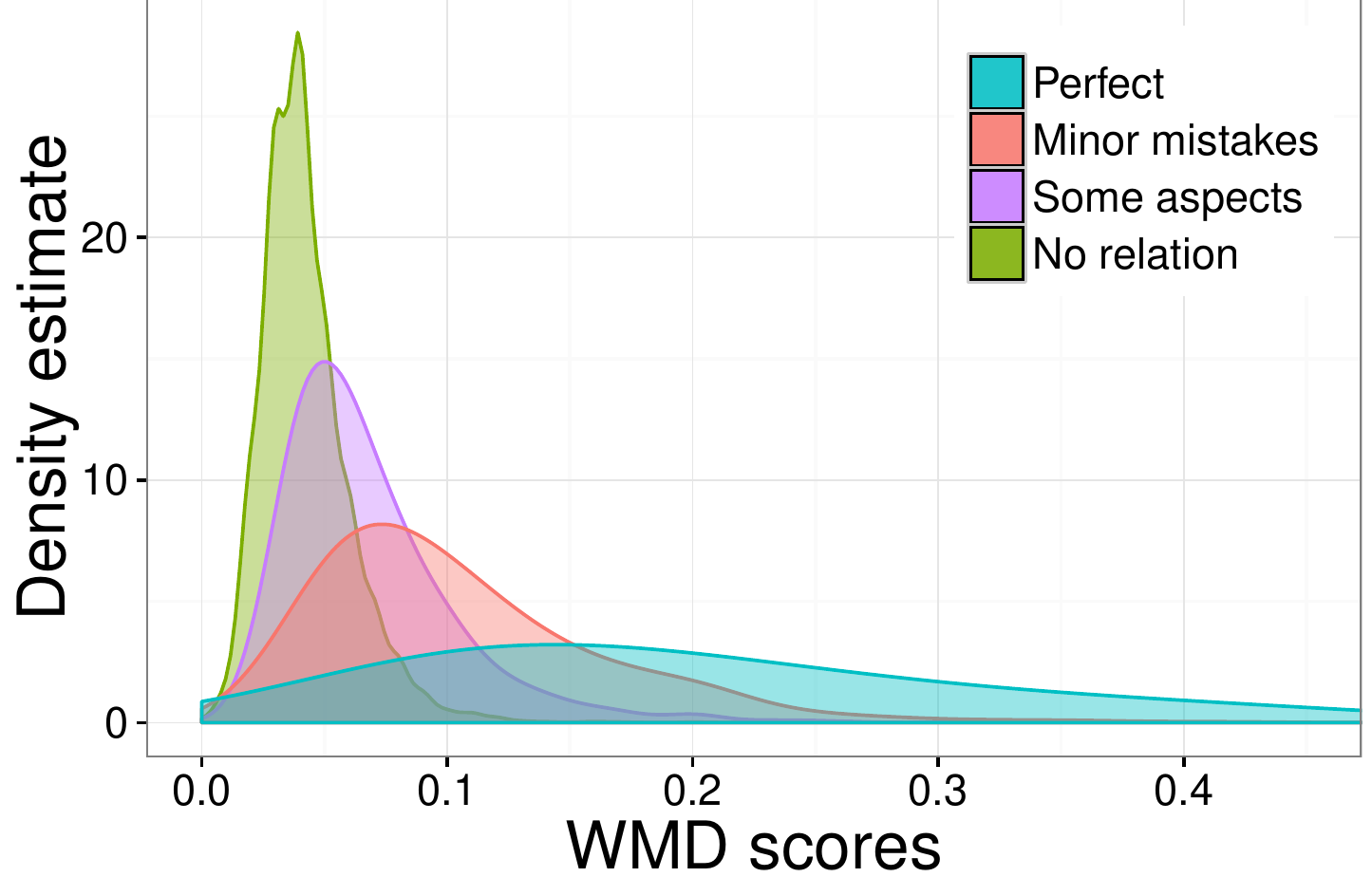}
\caption{Score distributions of the metrics on \textsc{flickr-8k} dataset. Four different rating scales are used: 1 for no relation, 2 for minor mistakes, 3 for some true aspects and 4 for perfect match. For \textsc{cide}r and \textsc{spice} metrics, square-root transform is performed on the $y$-axis to better illustrate how the score distributions overlap with each other.}
\label{fig:scoredist}
\end{figure*}

To analyze statistical significance in the automatic metrics listed in Section~\ref{sec:metrics}, we use the publicly available \textsc{flickr-8k}~\cite{elliott-keller:2014:P14-2} and \textsc{composite}~\cite{aditya2015images} datasets, which we describe below. We note that in our experiments, we first lowercase and tokenize the candidate and reference captions using \texttt{ptbtokenizer.py} script from \textsc{ms coco} evaluation tools\footnote{\url{https://github.com/peteanderson80/coco-caption}}. We use the implementations of the metrics from the same evaluation kit with the exception of \textsc{wmd}. For the \textsc{wmd} metric, we employ the code provided by Kusner et al.~\shortcite{kusner2015word}\footnote{\url{https://github.com/mkusner/wmd}}.

\textsc{flickr-8k}\footnote{\url{https://github.com/elliottd/compareImageDescriptionMeasures}} dataset contains quality judgements for 5822 candidate sentences for the images in its test set~\cite{hodosh2013framing}. These judgements are collected from 3 human experts and they are on a scale of $[1,4]$, with a score of 1 denoting a description totally unrelated to the image content, and 4 meaning a perfect description for the image. Candidate captions are all obtained from a retrieval based model, hence they are grammatically correct. 

\begin{figure*}[!t]
\centering
\begin{tabular}{ccc}
 \multirow{1}{*}[0.25\textwidth]{\rotatebox{90}{\textsc{flickr-8k}}} &
\includegraphics[width=0.3\textwidth]{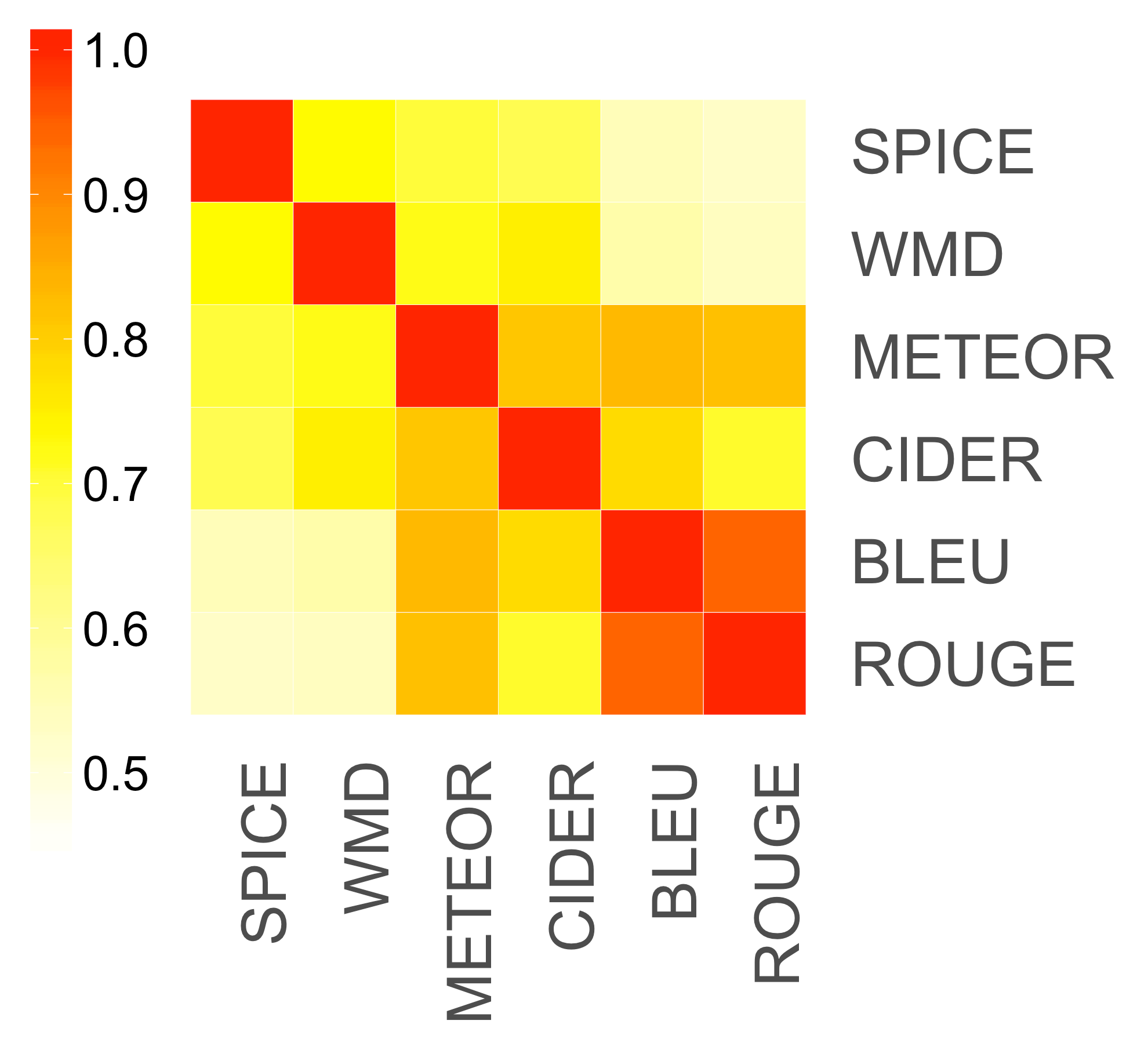} &
\includegraphics[width=0.3\textwidth]{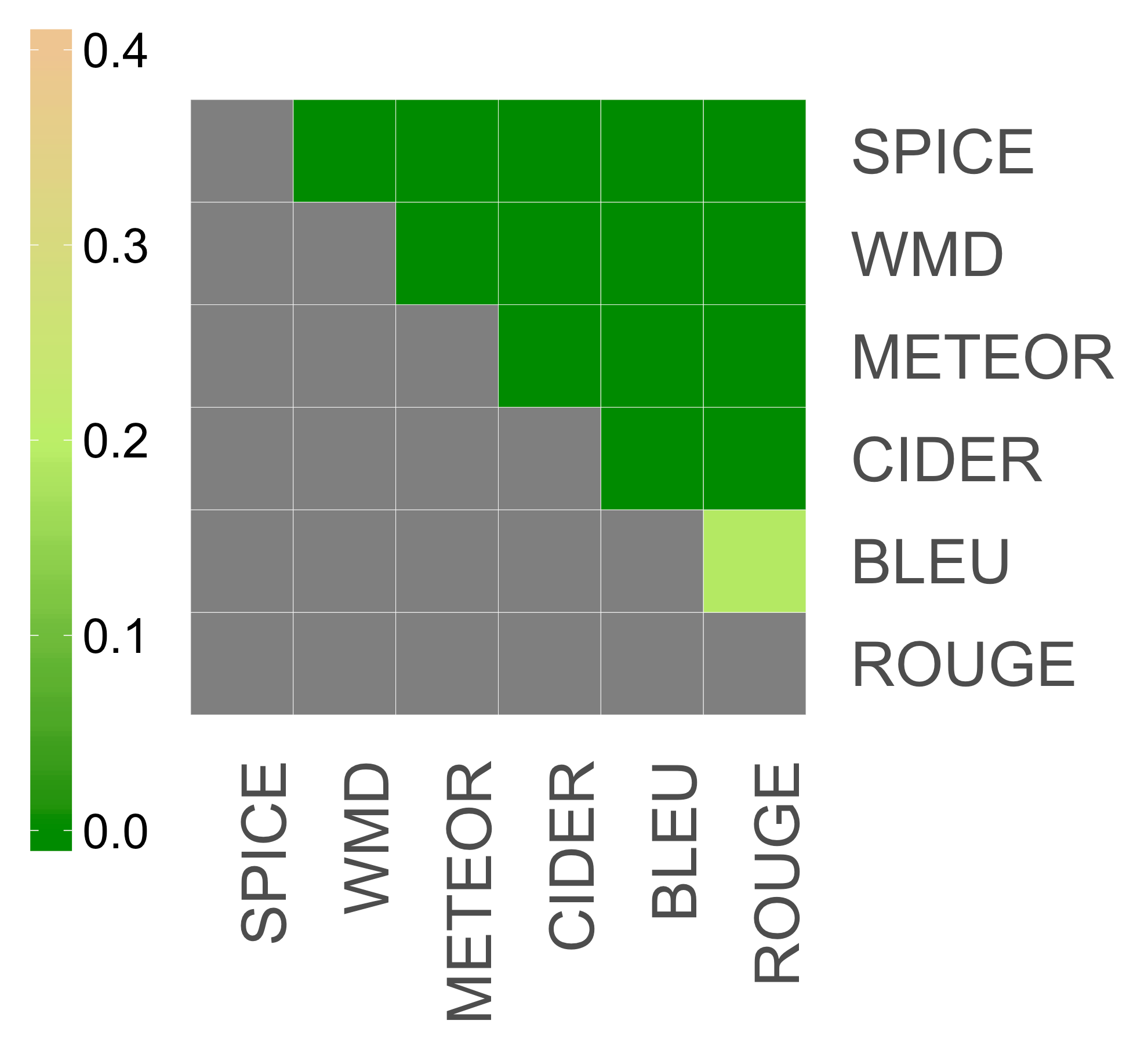} \\
\multirow{1}{*}[0.25\textwidth]{\rotatebox{90}{\textsc{composite}}}& \includegraphics[width=0.3\textwidth]{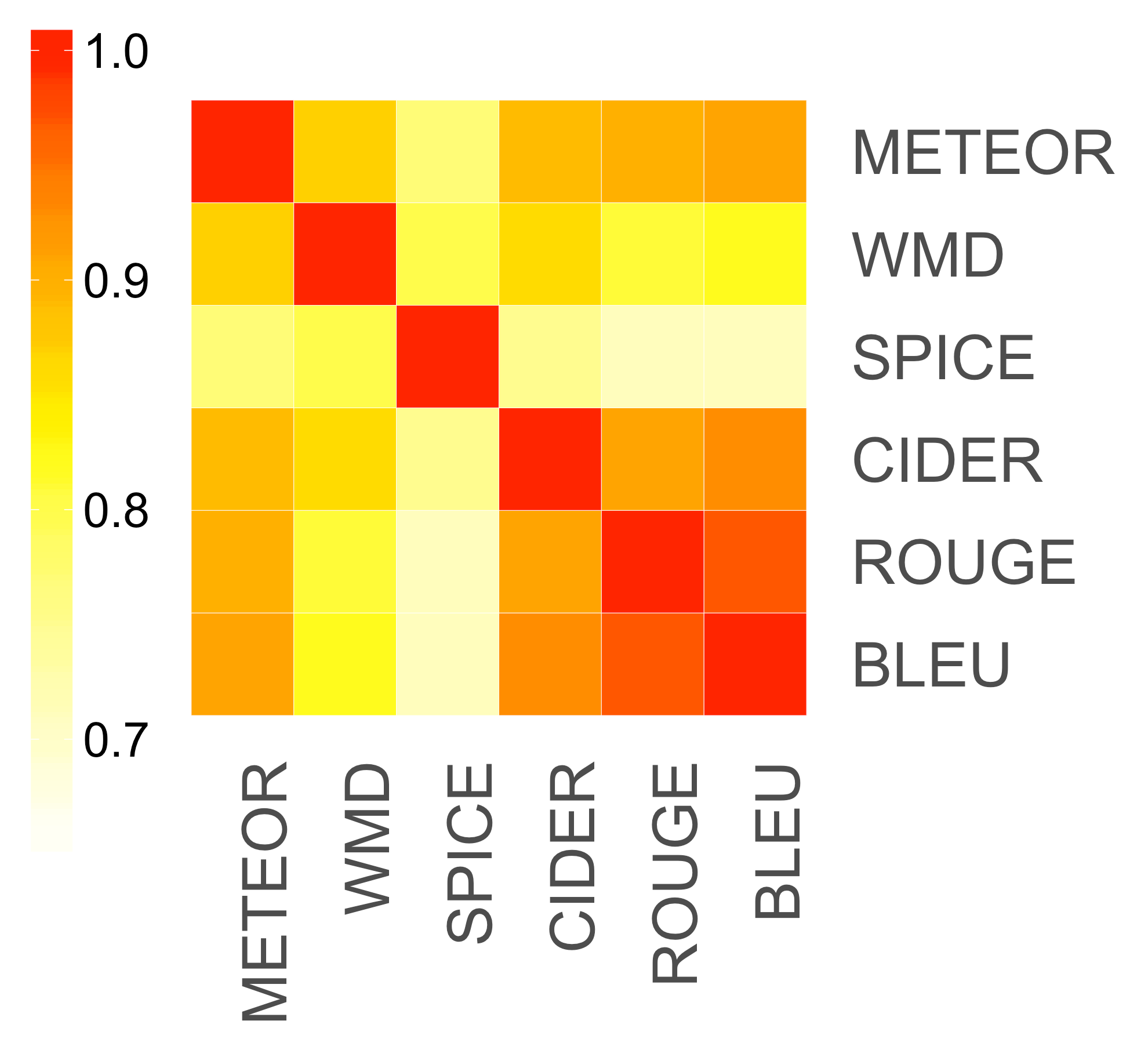} &
\includegraphics[width=0.3\textwidth]{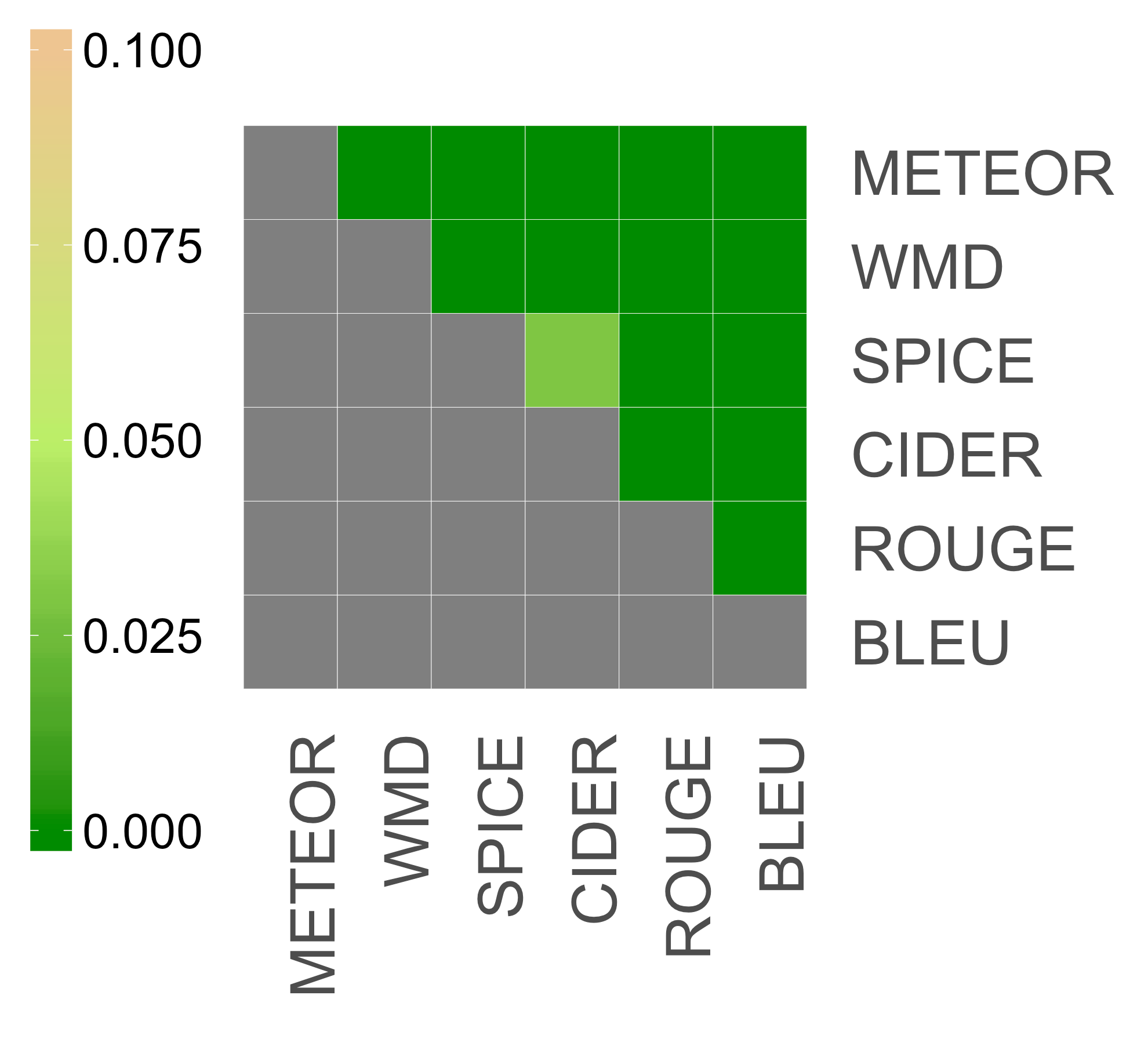}\\
& (a) Spearman's correlation & (b) Statistical significance 
\end{tabular}
\caption{Significance test results for pairs of automatic metrics on \textsc{flickr-8k} and \textsc{composite} datasets. (a) Spearman correlation between pairs of metrics; and (b) $p$-value of Williams significance tests, green cells indicate a significant win for metric in row i over metric in column j.}
\label{fig:correlations-williams}
\end{figure*}

\textsc{composite}\footnote{\url{https://imagesdg.wordpress.com/image-to-scene-description-graph}} dataset contains human judgements for 11,985 candidate captions for the subsets of \textsc{flickr-8k}~\cite{hodosh2013framing}, \textsc{flickr-30k}~\cite{young2014image} and \textsc{ms coco}~\cite{lin2014microsoft} datasets. The AMT workers were asked to judge the candidate caption for an image using two aspects: (i) \emph{correctness}, and (ii) \emph{thoroughness} of the candidate caption, both on a scale of $[1,5]$ where 1 means not relevant/less detailed and 5 denotes the candidate caption perfectly describing the image. Candidate captions were sampled from the human reference captions and the captioning models in~\cite{aditya2015images,karpathy2015deep}.

Table~\ref{tab:correlations} shows Pearson's, Spearman's and Kendall's correlation of the metrics with the human judgements in \textsc{flickr-8k} and \textsc{composite} datasets. For~\textsc{flickr-8k}, we follow the methodology in~\cite{elliott-keller:2014:P14-2} and compute correlations with the human expert scores. On the other hand. for~\textsc{composite}, we report the mean of the correlations with {\em correctness} and {\em thoroughness} scores. In terms of these correlations, while \textsc{spice} produces the highest quality comparisons in \textsc{flickr-8k}, \textsc{wmd} and \textsc{meteor} give better results in \textsc{composite} in general. However, if one further inspects the score distributions of the metrics (on \textsc{flickr-8k} dataset) shown in Figure~\ref{fig:scoredist}, while \textsc{spice} can identify irrelevant captions remarkably well, it can not effectively distinguish bad captions from relatively better ones.

In Figure~\ref{fig:correlations-williams}(a), we show Spearman's correlation between each pair of metrics, where the metrics are ordered from highest to lowest correlation with human judgements\footnote{Here, we only report Spearman's correlation since, compared to Pearson's, it provides a more consistent ranking of the metrics across the two datasets, and is similar to Kendall's correlation.}. Overall, the pairwise correlations are generally high for both datasets. We additionally observe that the metrics which depend on similar structures are grouped together using these correlations. For example, the $n$-gram based metrics \textsc{bleu} and \textsc{rouge} provide scores that are highly correlated with each other for \mbox{\textsc{flickr-8k}}. The correlations within \textsc{composite} dataset are even very high for all the metrics that consider $n$-grams, namely \textsc{bleu}, \textsc{cide}r, \textsc{meteor} and \textsc{rouge}. On the other hand, the correlations of these metrics against \textsc{spice} and \textsc{wmd} are not that high. Moreover , the pairwise correlations between \textsc{spice} and \textsc{wmd} are relatively low as well. All these findings suggest that these three groups of metrics, the $n$-gram based metrics, the scene-graph based \textsc{spice} and the word embedding based \textsc{wmd}, can be complementary to each other.

Finally, in Figure~\ref{fig:correlations-williams}(b), we provide the results of Williams significance test, which compares two different metrics with respect to their correlations against human judgements. Our results show that all the metric pairs have a significant difference in correlation with human judgement at $p < 0.05$. This reveals that the pair of metrics which has close correlation scores with human judgements (e.g. \textsc{spice} and \textsc{wmd} in \textsc{flickr-8k} dataset) are found to be statistically different than each other. These findings collectively support our previous conclusion that all metrics considered here can complement each other in evaluating the quality of the generated captions.

\subsection{Accuracy}
\label{ssec:accuract}

In this section, following the methodology introduced in~\cite{vedantam2015cider}, we analyze the ability of each metric to discriminate certain pair of captions from one another in reference to a groundtruth caption. We employ the human consensus scores while evaluating the accuracies. In particular, for evaluation, a triplet of descriptions, one reference and two candidate descriptions, is shown to human subjects and they are asked to determine the candidate description that is more similar to the reference. A metric is accurate if it provides a higher score to the description chosen by the human subject as being more similar to the reference caption. For this analysis, we carry out our experiments on \textsc{pascal-50s} and \textsc{abstract-50s} datasets\footnote{\url{http://vrama91.github.io/cider}}. We consider different kinds of pairs such as (human-human correct) HC, (human-human incorrect) HI, (human-machine) HM, and (machine-machine) MM. As the candidate sentences are generated by both humans and machines, each test scenario has a different level of difficulty.
 
\begin{table*}[!t]
\centering
\caption{Description-level classification accuracies of automatic evaluation metrics.}
\begin{tabular}{l@{\hspace{0.05\linewidth}}rrr@{\hspace{0.075\linewidth}}rrrrr} \toprule
& \multicolumn{3}{l}{\textsc{abstract-50s}} & \multicolumn{5}{c}{\textsc{pascal-50s}} \\ 
 & HC & HI & Avg. & HC & HI & HM & MM & Avg. \\\midrule  
 
\textsc{wmd} & 0.65	& 0.93 & 0.79 & \textbf{0.71}	& \textbf{0.99} & 	0.93 & 	\textbf{0.74} & 	\textbf{0.84} \\  
\textsc{spice}    & 0.62	 & 0.89 & 	0.76 & 0.66	& 0.98 & 0.85 & 0.72 & 	0.81 \\  
\textsc{cide}r    & \textbf{0.76}	& \textbf{0.95} & 	\textbf{0.86} & 0.69	& \textbf{0.99} & 	\textbf{0.94} & 	0.66 & 	0.82 \\ 
\textsc{meteor}   & 0.60	 & 0.90 & 	0.75 & 0.69	 & \textbf{0.99} & 	0.90 & 	0.65 & 	0.81 \\
\textsc{bleu}     & 0.69	 & 0.89 & 	0.79 & 0.67	 & 0.97 &	\textbf{0.94} & 	0.60 & 	0.80 \\ 
\textsc{rouge}    & 0.65	 & 0.89 & 	0.77 & 0.68	 & 0.97 &	0.92 & 	0.60 & 	0.79 \\ 
\bottomrule 
\end{tabular} 
\label{tab:accuracy}
\end{table*}

\begin{figure*}[!t]
\centering
\includegraphics[width=\textwidth]{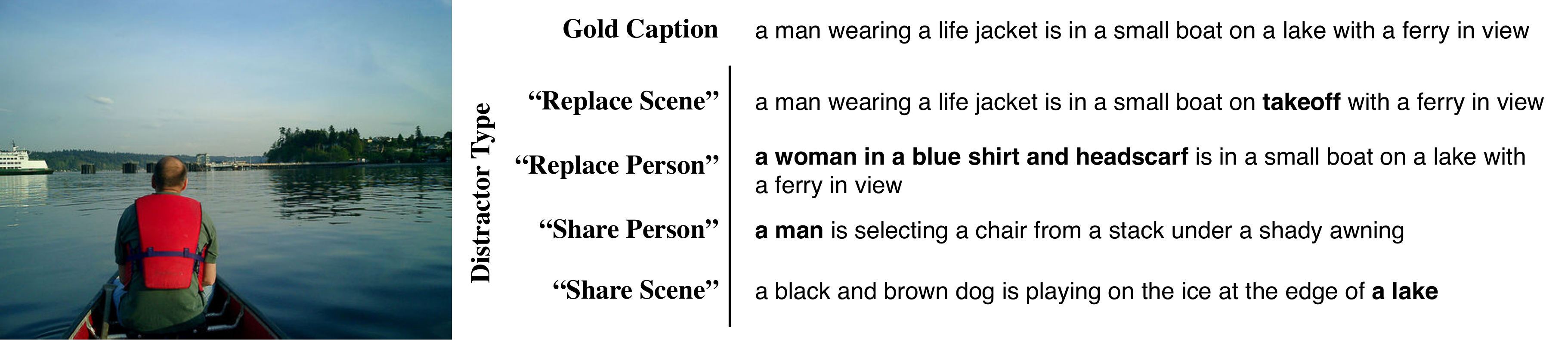}
\caption{Distracted versions of an image descriptions for a sample image.}
\label{fig:distractions}
\end{figure*}

{\textsc{abstract-50s}}~\cite{vedantam2015cider} dataset is a subset of the Abstract Scenes Dataset~\cite{zitnick2013bringing}, which includes 500 images containing clipart objects in everyday scenes. Each image is annotated with 50 different descriptions. For evaluation, 48 of these 50 descriptions are used as reference descriptions and the remaining 2 descriptions are employed as candidate descriptions. For 400 pairs of these descriptions, human consensus scores are available, with the first 200 are for HC and the remaining 200 are for HI. 

{\textsc{pascal-50s}}~\cite{vedantam2015cider} dataset is an extended version of the Pascal Sentences~\cite{farhadi2010every} dataset that contains 1000 images from PASCAL Object Detection challenge~\cite{everingham2010pascal} of 20 different object classes like person, car, horse, etc. This  version includes 50 captions per image and human judgements for 4000 candidate pairs for the aforementioned binary-forced choice task, which are all collected through Amazon Mechanical Turk (AMT). For this dataset, all four different categories are available, having 1000 pairs for each category. 

\begin{table*}[!t]
\vspace{-0.25cm}
  \centering
  \caption{Distraction analysis.}
  \resizebox{0.8\linewidth}{!} {
  \begin{tabular}{lrrrrrrr}
  \toprule 
Case & \# Instances & \textsc{bleu} & \textsc{meteor} & \textsc{rouge} & \textsc{cide}r & \textsc{spice} & \textsc{wmd} \\ \midrule 
Replace-Scene & 2514  & 0.62	&  0.69 & 0.63 &	\textbf{0.83} &	0.54 &	0.76 \\ 
Replace-Person & 5817 & 0.73	&  0.77	& 0.78	&  0.78 & 	0.67 & 	\textbf{0.80} \\ 
Share-Scene    & 2621 & 0.79	&  0.85	& 0.79	&  0.81 &	0.70 &	\textbf{0.87} \\ 
Share-Person   & 4596 & 0.78	&  0.85	& 0.78	&  0.83	&   0.67 &	\textbf{0.88} \\ \midrule
Overall        & 15548 & 0.73   &  0.79 & 0.75  &  0.81 &   0.65 &  \textbf{0.83}         \\ 
 \bottomrule 
  \end{tabular}}
   \label{tab:distraction}
\end{table*} 

In Table~\ref{tab:accuracy}, we present caption-level classification accuracy scores of automatic evaluation metrics at matching human consensus scores. On \textsc{abstract-50s} dataset, the \textsc{cide}r metric outperforms all other metrics in both HC and HI cases. On the other hand, on \textsc{pascal-50s} dataset, the \textsc{wmd} metric gives the best scores in three out of four cases. Especially, it is the most accurate metric at matching human judgements on the challenging MM and HC cases, which require distinguishing fine-grained differences between descriptions. On average, the performances of all the other metrics are very similar to each other.

\subsection{Robustness}
\label{ssec:robustness}

In this section, we evaluate the robustness of the automatic image captioning metrics. For this purpose, we employ the binary (two-alternative) forced choice task introduced in \cite{hockenmaier-acl-ws} to compare the existing image captioning models. For a given image, this task involves distinguishing a correct description from its slightly distracted incorrect versions. In our case, a robust image captioning metric should always choose the correct caption over the distracted ones.

In our experiments, we use the data\footnote{\url{http://nlp.cs.illinois.edu/HockenmaierGroup/Papers/VL2016/HodoshHockenmaier16_BinaryTasks_Data.tar}} provided by the authors for a subset of \textsc{flickr-30k}~\cite{hodosh2013framing}. Specifically, we consider four different types of distractions for the image descriptions, namely 1) Replace-Scene, 2) Replace-Person, 3) Share-Scene, and 4) Share-Person, which results 15548 correct and distracted caption pairs in total. For Replace-Scene and Replace-Person tasks, the distracted descriptions were artificially constructed by replacing the main actor (first person) and the scene in the original caption by random person and scene elements, respectively. For Share-Scene and Share-Person tasks, the distracted captions were selected from the sentences from the training part of \mbox{\textsc{flickr-30k}}~\cite{young2014image} dataset whose actor or scene chunks share the similar main actor or scene elements with the correct description. Figure~\ref{fig:distractions} presents an example image together with the original description and its distracted versions.

We compare each correct caption available for an image with the remaining correct and distracted captions for that image by considering tested evaluation metrics, and then estimate an average accuracy score. In Table~\ref{tab:distraction}, we present the classification accuracies of the evaluation metrics for each distraction type. As can be seen, the \textsc{wmd} metric gives the best results for three out of four categories, and provides the second best result for the Replace-Scene case. Overall, \textsc{meteor} and \textsc{cide}r metrics seem to be also robust to these distractions. The very recently proposed \textsc{spice} metric performs the worst for this task. This is somewhat expected as it is even affected by the use of synonyms of the words as we have previously shown in Table~\ref{tab:hand-distractor}.

\subsection{Discussion}
\label{ssec:discussion}

As the experiments on quality, accuracy and robustness tests demonstrate in Sections~\ref{ssec:williams}-\ref{ssec:robustness}, existing automatic image captioning metrics all have some strengths and weaknesses due to their design choices. For example, while \textsc{spice}, \textsc{meteor} and \textsc{wmd} give the best performances in terms of our correlation analysis against human judgements, \textsc{cide}r and \textsc{wmd} provide the best classification scores for our accuracy experiments. Moreover, \textsc{cide}r, \textsc{meteor} and \textsc{wmd} are found to be less affected by the distractors. Overall, our analysis suggests that the recently proposed \textsc{wmd} document metric is also quite effective for image captioning since it has high correlations with the human scores, is much less sensitive to synonym swapping and additionally performs well at the accuracy and distraction tasks.

Our analysis also shows that the existing metrics both theoretically and empirically differ from each other with significant differences. Compared to the recent results of significance testing of machine translation and summarization metrics~\cite{graham-baldwin:2014:EMNLP2014,grahametal:15,graham2016accurate,graham:2015:EMNLP}, our results suggest that there remains much room for improvement in developing more effective image captioning evaluation metrics. We leave this for future work, but a very naive idea would be combining different metrics into a unified metric and we simply test this idea using score combination, after normalizing the score of each metric to the range $[0,1]$. Among all possible combinations, we find that the combination of \textsc{wmd}+\textsc{spice}+\textsc{meteor} performs the best with a Spearman's correlation of 0.66 for \textsc{flickr-8k} and 0.45 for \textsc{composite} dataset, yielding an improvement from \textsc{spice} (0.64 and 0.42). In addition, we should add that this unified metric significantly outperforms the individual metrics according to Williams test ($p<0.01$).

\section{Conclusion}
\label{sec:conclusion}

In this paper, we provide a careful evaluation of the automatic image captioning metrics, and propose to use \textsc{wmd}, which utilizes {\em word2vec} embeddings of the words to compute a semantic similarity of sentences. We highlight the drawbacks of the existing metrics, and we empirically show that they are significantly different than each other. We hope that this work motivates further research into developing better evaluation metrics, probably learning based ones, as previously studied in machine translation literature~\cite{kotani2010machine,guzman2015-ACL}. We also observe that incorporating visual information (via Scene-graph used by \textsc{spice}) and semantic information (via \textsc{wmd}) is useful for the caption evaluation task, which motivates the use of multimodal embeddings~\cite{kottur2015visual}. 

\section*{Acknowledgments}
We thank the anonymous reviewers for their valuable comments. This work is supported in part by The Scientific and Technological Research Council of Turkey (TUBITAK), with award no 113E116.

\bibliography{eacl2017}
\bibliographystyle{eacl2017}
\end{document}